\def\year{2019}\relax
\documentclass[letterpaper]{article}
\usepackage{aaai19}
\usepackage{times}
\usepackage{helvet}
\usepackage{courier}
\usepackage{url}
\usepackage{graphicx}
\usepackage{microtype}
\usepackage{multirow}
\usepackage{amsfonts}
\usepackage{multirow}
\usepackage{amsmath}
\usepackage{amssymb}
\usepackage{enumitem}
\usepackage{xcolor}
\usepackage{booktabs} 
\usepackage{url}
\usepackage{latexsym}
\usepackage{graphicx}
\usepackage{array}
\usepackage{xcolor}
\usepackage{color, colortbl}
\definecolor{LGray}{gray}{0.9}
\definecolor{Gray}{gray}{0.8}
\definecolor{DGray}{gray}{0.7}

\usepackage{tikzpagenodes}

\makeatletter
\newcommand*\titleheader[1]{\gdef\@titleheader{#1}}
\AtBeginDocument{%
  \let\st@red@title\@title
  \def\@title{%
    \bgroup\normalfont\large\centering\@titleheader\par\egroup
    \vskip1.5em\st@red@title}
}
\makeatother

\makeatother
\newcolumntype{L}[1]{>{\raggedright\let\newline\\\arraybackslash\hspace{0pt}}m{#1}}
\newcolumntype{C}[1]{>{\centering\let\newline\\\arraybackslash\hspace{0pt}}m{#1}}
\newcolumntype{R}[1]{>{\raggedleft\let\newline\\\arraybackslash\hspace{0pt}}m{#1}}
\begin{document}
 
%
\title{Understanding and Measuring Psychological Stress Using Social Media}
\author{Sharath Chandra Guntuku,$^{1}$ Anneke Buffone,$^{1}$ Kokil Jaidka$^{2}$ \\ {\bf \Large Johannes C. Eichstaedt,$^{1}$ Lyle H. Ungar$^{1}$}\\
       $^{1}$University of Pennsylvania, $^{2}$Nanyang Technological University\\
       \{sharathg@sas, annekeb@sas, jeich@sas, ungar@cis\}.upenn.edu,jaidka@ntu.edu.sg
       }

\maketitle

\begin{abstract}
A body of literature has demonstrated that users' mental health conditions, such as depression and anxiety, can be predicted from their social media language. There is still a gap in the scientific understanding of how psychological stress is expressed on social media. Stress is one of the primary underlying causes and correlates of chronic physical illnesses and mental health conditions. In this paper, we explore the language of psychological stress with a dataset of 601 social media users, who answered the Perceived Stress Scale questionnaire and also consented to share their Facebook and Twitter data. Firstly, we find that stressed users post about exhaustion, losing control, increased self-focus and physical pain as compared to posts about breakfast, family-time, and travel by users who are not stressed. Secondly, we find that Facebook language is more predictive of stress than Twitter language. Thirdly, we demonstrate how the language based models thus developed can be adapted and be scaled to measure county-level trends. Since county-level language is easily available on Twitter using the Streaming API, we explore multiple domain adaptation algorithms to adapt user-level Facebook models to Twitter language. We find that domain-adapted and scaled social media-based measurements of stress outperform sociodemographic variables (age, gender, race, education, and income), against ground-truth survey-based stress measurements, both at the user- and the county-level in the U.S. Twitter language that scores higher in stress is also predictive of poorer health, less access to facilities and lower socioeconomic status in counties. We conclude with a discussion of the implications of using social media as a new tool for monitoring stress levels of both individuals and counties.
\end{abstract}


\section{Introduction}
\begin{figure}[!ht]
	\centering
	\includegraphics[width=1\columnwidth]{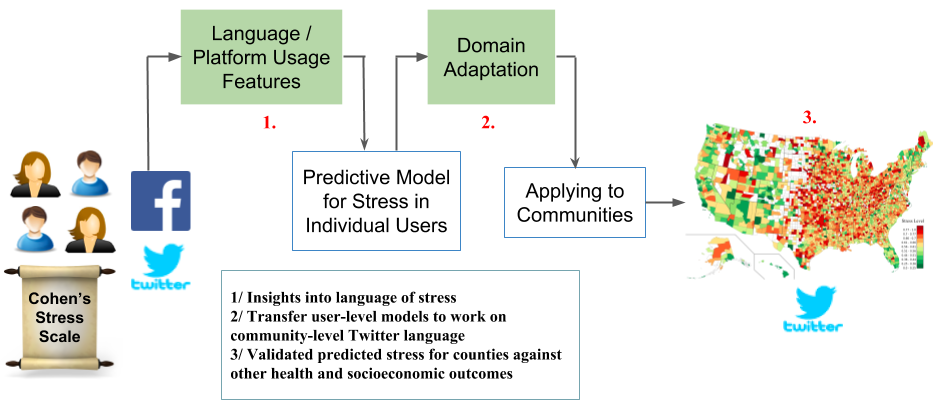}
	\caption{Overview of the approach taken to build language based stress prediction model.}
	\label{fig:approach}
\end{figure}

Stress is defined as perceived distress caused by an interaction between a person and their environment \cite{cohen2001psychological}. While people can handle stress better or worse depending on their general coping skills, experiencing stress too frequently is known to affect well-being and physical and mental health negatively. Stress is seen to be a single pervasive trait influencing health, through a broad range of negative affective states and somatic pathways \cite{mcewen1993stress}. Considering that the symptoms associated with depression and other severe mental health conditions are very severe, alleviating psychological stress and promoting a healthy lifestyle is more efficient compared to treating a more acute and chronic condition \cite{wilkinson2005impact}. 

People are increasingly using social media platforms in order to inform others about their mental states, solicit social support, as well as keep records of their daily activities, preferences, and interests. Notwithstanding the challenge of working with a non-random, non-representative sample of social media users, studies have identified the markers of self-disclosure which concern depression \cite{guntuku2017detecting}, schizophrenia \cite{ernala2017linguistic}, ADHD \cite{guntuku2017language}, alcohol consumption \cite{liu2017assessing}, and personality \cite{guntuku2017studying}. With respect to stress, the linguistic features of event-related stress have been predicted from social media posts about experiences such as travel and work \cite{lin2014user}; however, these findings cannot be applied to improve the psychological understanding of stress, because people suffering from chronic stress do so irrespective of stressful events. For instance, preparing for an exam is a stressful event, while chronically feeling overwhelmed with responsibilities is trait-related stress. Another research gap is that the previous work has focused on known stressors collected using search keywords \cite{thelwall2017tensistrength}. However the labels thus acquired likely have personality confounds \cite{preotiuc2015role}, emphasizing the need for using stronger ground truth. Instead, we anticipate that insights into psychological stress could help in (a) designing social-media-based interventions to enable a low-stress lifestyle, and (b) developing a better understanding of regional variations in stress.


The ubiquitous nature of smart devices and Internet access in almost all parts of the world means that social media is a potentially powerful tool to measure the psychological states and behaviors of people at both micro- (individual) and macro- (county) levels. 
However, except for a few studies, little has been done to explore how to scale language models to study regions, and no work yet has attempted to do this for stress. Although some studies have analyzed the geographic variation in social media language corresponding to chronic illnesses \cite{culotta2014estimating}, depression \cite{bagroy2017social}, well-being \cite{schwartz2013characterizing}, heart-disease \cite{eichstaedt2015psychological}, and happiness \cite{quercia2012tracking}, they are often challenged by (a) a limited understanding of how to scale user-level models to measure counties, and (b) the lack of ground truth about a large population. Recent work has shown the need to use weighting and scaling techniques to transform user-level language estimates from Twitter to county-level estimates \cite{rieman2017domain}, in order to avoid the ecological fallacies reported in the studies mentioned above. However, there are challenges associated with transferring predictive models from one social media platform to another, because of differences in self-disclosure on Facebook vs. Twitter \cite{kokil2018}. This study contributes with (a) effective off-the-shelf methods for cross-domain adaptation of user-level stress models to measure county-level stress, and (b) validation against ground-truth survey data built on over two million responses. 


\begin{table}[t!]
	\centering
	\small
		\caption{Items on the Cohen's stress scale: Each question is assessed on a Likert Scale. (-) indicates reverse coded items.}
	
		\resizebox{\columnwidth}{!}{
	\begin{tabular}{l}
		\hline
		In the last month, how often have you: \\ \hline
		- been upset because of something that happened unexpectedly?\\
		- felt that you were unable to control the important things in your life?\\ 
		- felt nervous and ''stressed''?\\ 
		- felt confident about your ability to handle your personal problems? (-)\\
		- felt that things were going your way? (-)\\ 
		- found that you could not cope with all the things that you had to do?\\ 
		- been able to control irritations in your life? (-)\\
		- felt that you were on top of things?(-)\\
		- been angered because of things that were outside of your control?\\
		- felt difficulties were piling up so high that you could not overcome them? \\ \hline
	\end{tabular}
}
\label{table:stress}
\end{table}

To summarize, the research gaps in previous work are the lack of language models to predict psychological stress, a limited understanding of scaling and transforming Facebook language models to work on Twitter, and the lack of validation against the region-level ground truth. We show in Figure \ref{fig:approach}, how our study uses transfer learning to adapt user-level models trained on their Facebook language, to predict county-level stress from the county-level Twitter language. This is necessary because it is easier to train predictive models on the language of a small population of social media users, but it is expensive to survey entire counties for training purposes. Furthermore, county-level social media language is only available for Twitter, where approximately 20\% of all public posts are geo-tagged with their location information, and they can be easily mined by using Twitter's Streaming API\footnote{https://developer.twitter.com/en}. On the contrary, Facebook needs user authentication for accessing their posts which is resource intensive to collect from a large number of people for this research study. Adapting models trained at user-level to predict stress in counties has multiple applications in monitoring health and well-being in counties, especially where survey data is hard to collect. Thus motivated, in this paper we address the following research questions:
\begin{enumerate}[label={\textbf{RQ\arabic*}},leftmargin = *]
	\item How does social media language of users who are stressed differ from those who are not?
	\item How do Facebook and Twitter language differ in predicting user-level stress? Since county-level language is easily available on Twitter, can off-the-shelf domain adaptation algorithms be used to improve prediction on Twitter language?
	\item How do domain-adapted social media-based measurements of stress at the county level correlate with health behaviors and socioeconomic characteristics at the county level?
\end{enumerate}

\section{RQ1: Differential Language Analysis of Stressed Users}

\subsection{Methods}

\subsubsection{User-level social media data:} 
We deployed a survey on Qualtrics\footnote{www.qualtrics.com/Survey-Software} (a platform similar to Amazon Mechanical Turk), comprising several demographic questions (age, gender, race, education, and income) and the Cohen's 10-item Stress scale \cite{cohen1997measuring} (Table \ref{table:stress}), and invited users to share access to their Facebook status updates and/or Twitter usernames.  Users received an incentive for their participation, and we obtained their informed consent to access their Facebook and Twitter posts. All users were based in the US. This study received approval from the IRB of our institution.

Out of all users who took the survey, 601 users completed the survey and had active accounts with more than 900 words on both Facebook and Twitter. We collected their Facebook posts by using the Facebook Graph API and downloaded their Twitter posts using the Twitter API. Of these 601 users, 265 self-identified as female. The mean age of the sample was 38. The stress scores range from 6 to 39 (mean 30). Each item in the scale is scored on 0-4, with an absolute maximum summing to 40. 

\begin{figure*}[t!]
\resizebox{\textwidth}{!}{
\begin{tabular}{cc}
  \includegraphics[width=0.3\columnwidth]{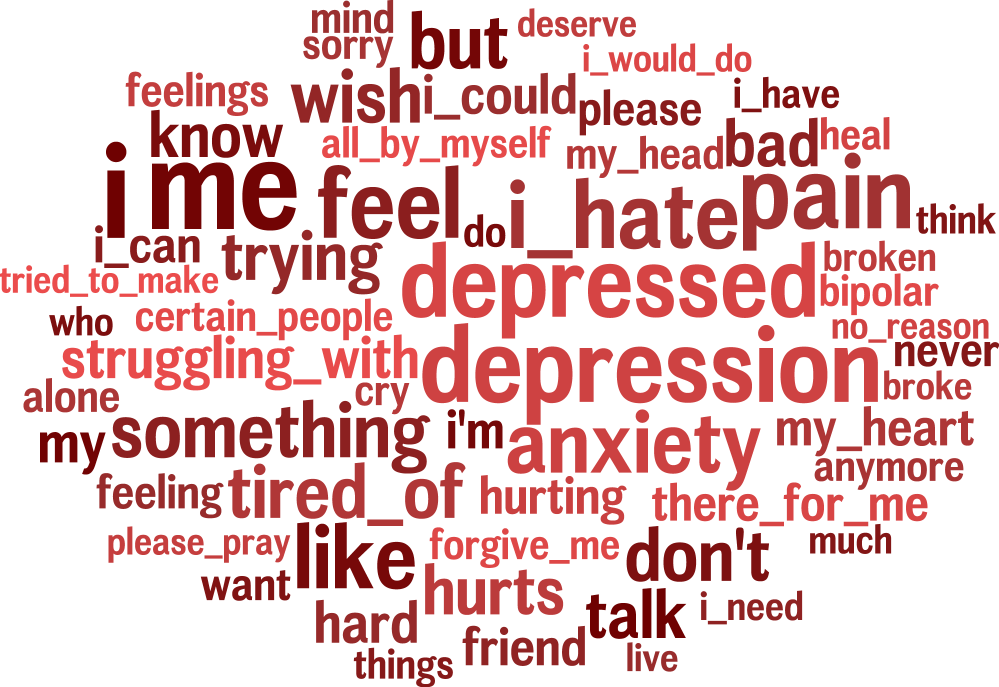} &   \includegraphics[width=0.3\columnwidth]{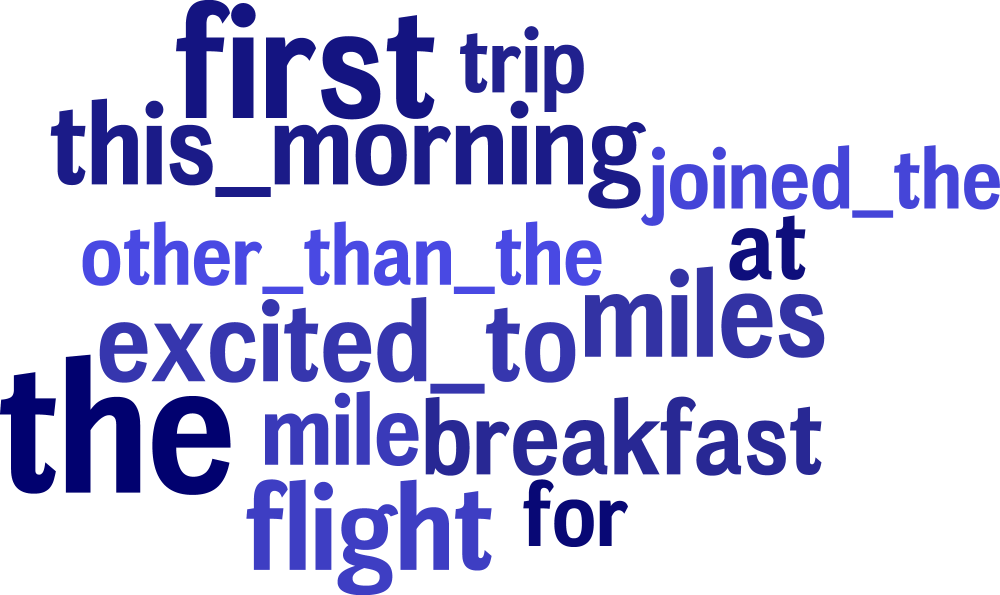} \\
\end{tabular}
}
\caption{Words and phrases associated with a) high-stress users (red) and b) low-stress users (blue). The size of the word indicates the correlation strength and the color indicates frequency (darker is more frequent). Correlations are controlled for age and gender, and are significant at $p<.01$, two-tailed t-test Bonferroni p-correction.}
\label{fig:stress2d}
\end{figure*}

\subsubsection{Features:} We process all our social media posts using the HappierFunTokenizer available with the DLATK package \cite{schwartz2017dlatk} which is emoticon- and social media-aware. We then represent the language of each user and county as a set of features. In the dictionary-based method, we transform social media language into numerical features representing percentage proportions of lexical categories in an existing dictionary. In the data-driven method, we transform language into numerical features which represent the proportions of word clusters which are statistically similar according to their frequency distributions. 

\textbf{LIWC:} We use Linguistic Inquiry and Word Count (LIWC) \cite{pennebaker2007linguistic}, a dictionary comprising 73 different psycholinguistic categories (e.g., topical categories, emotions, parts-of-speech) to represent the language of each user and county as the normalized frequency distributions. 

\textbf{Topics:} In the data-driven method, we represent the language of each user and county as normalized frequency distributions for a set of topics derived using Latent Dirichlet Allocation. These topics are an open-source resource available through the DLATK package \cite{schwartz2017dlatk} and were trained on a corpus of over 20 million Facebook statuses \cite{schwartz2013personality}.

\textbf{TensiStrength:} To detect direct and indirect expressions of stress or relaxation, a stress lexicon \cite{thelwall2017tensistrength} is used. We obtain stress scores at the sentence level, from each post on Facebook and Twitter and aggregate them to users by calculating mean stress scores. 

\textbf{User engagement} We extracted features such as the number of posts made between 12am-6am,  mean message length, and number of URLs \& hashtags. These features were shown to be predictive of stress by \cite{lin2014user}.

\subsection{Results}

We identify the linguistic characteristics indicative of high and low psychological stress based on the social media language of individual users on Facebook. We conducted a similar analysis on Twitter and present them in the appendix. Since we explore several features simultaneously, we consider coefficients significant if they are less than a Bonferroni-corrected two-tailed p of 0.01 (i.e., when examining 1000 features, in the case of words and phrases, a passing p-value is less than $1 x 10^{-5}$ ; when examining 2000 topics p-value is less than $5 x 10^{-6}$, and when examining 73 LIWC categories p-value is less than $1.3 x 10^{-4}$).

\textbf{Ngrams:} We extract 1-,2-, and 3-grams from all posts to analyze significant associations between words \& phrases and stress. Figure \ref{fig:stress2d} provides a visualization of the Pearson correlations of words and phrases with stress on Facebook. In general, the language of stress has a prominent self-focus. Given that psychologists have found stress to be a state resulting from an individual assessing the demands of the stressful stimulus and determining own coping abilities~\cite{staal2004stress}, this result is very intuitive. It should be noted that increased self-focus in stressful situations is likely adaptive, but a prolonged self-focus in one's thoughts, especially in the context of negative affect, is known to psychologists as rumination and linked to negative effects for health and well-being~\cite{moberly2008ruminative}. Hate could signify aggressive or angry affect, which also tends to be maladaptive and could signify the experience of frustration while perceiving resources fall short of perceived demands. 

Words and phrases such as ``me'', ''I had'', ``feel like'', ''I don't'' and ''I hate'' are significantly correlated with high stress. Furthermore, the language of high stress appears to be marked by expressions of perceived lack of control and expressions of a need state or lack of resources (``struggling with'', ``tired of'', ``I need''), as well as negative-angry affect (``hate'', ``I hate''). Further, high stress language seem to be comorbid with mental health conditions (``depression'', ``depressed'', ``anxiety'', ``bipolar''). It is interesting that language reflects the adverse effects that stress can have on health \cite{watson1989health}. The language of low stress has prominent positive affect (``excited to''), discussions of meals (``breakfast''), as well as feelings of social inclusion (``joined the''). The language of low stress comprises the discussion of meals, specifically ``breakfast'', which may indicate relaxation and enjoyment inasmuch as meals are often taken in the company of others, social inclusion, as well as with positive affect (``excited to''). 

\begin{table}[!ht]
	\renewcommand{\arraystretch}{1.1}
	\setlength\tabcolsep{0.02in}
	\centering
	\footnotesize
		\caption{\small Pearson correlations (top 5) between stress score and LIWC features extracted on Facebook, when controlling for age and gender. All features are significant at $p<.01$, two-tailed t-test and Bonferroni corrected.}
	
	\footnotesize
		\begin{tabular}{ >{\centering\arraybackslash}m{0.3\columnwidth}|>{\arraybackslash}m{0.4\columnwidth}|c}
		\hline
			\rowcolor{DGray}
			
			\rowcolor{Gray} \multicolumn{3}{l}{\textbf{LIWC}}   \\
			\hline    
			\textbf{Positively correlated} & most frequent words & r  \\
			\hline    
			1st Person Singular & I        & .22 \\
			Adverbs              &  very, really    & .17 \\
			Negation          & no, not, never    & .16 \\
			Negative Emotion  & hurt, sadness, anger & .17 \\
			Fillers           & I mean, you know     & .15 \\
			\hline
			\textbf{Negatively correlated} & most frequent words & r  \\
			\hline    
			1st Person Plural  & we         & -.18 \\
			Affiliation    & friend, social  & -.17 \\
			Positive Emotion       & love, nice, sweet & -.12 \\
			\hline
		\end{tabular}
\

\label{t:liwc}
\end{table}

\textbf{LIWC:} LIWC categories from Facebook language that significantly correlate with stress are shown in Table \ref{t:liwc}. The top correlated categories in Facebook comprise first person singular pronouns (``I''), indicating increased self-focus, which corroborate previous findings  \cite{pennebaker2002language} that self-references by individuals increase in emotionally vulnerable situations. Users on Facebook are more likely to use adverbs such as ``very'' and ``really'' to emphasize their point, and more likely to explicitly use words denoting negative emotions, such as ``hurt'' and ``anger''. Mentions of negative emotion confirms our expectations, because stress is an aversive state \cite{cohen1997measuring}. Filler word associations may indicate a lack of self-esteem or feeling down on oneself, and hedging as a result \cite{abouserie1994sources}. Negation further suggests a 'lack of' things, mentioned and experienced by those who are also prone to high stress. On the other hand, the first person plural pronouns (``We''), and the LIWC ``Affiliation'' categories are negatively correlated with stress, which implies that a high-stress individual often depicts themselves as isolated, with a certain disassociation from their social circles. Negative correlations of stress with positive emotion reflects that those individuals reporting lower stress are significantly more likely to express positive emotion.

\begin{table}[!ht]
	\renewcommand{\arraystretch}{1.1}
	\setlength\tabcolsep{0.02in}
	\centering
	\footnotesize
		\caption{\small Pearson correlations between stress score and Topics extracted from Facebook, when controlling for age and gender. Topic labels are manually created. All features are significant at $p<.01$, two tailed t-test and Bonferroni corrected. The top 5 topics are shown for both positive and negative correlations. }
	\footnotesize
		\begin{tabular}{ >{\centering\arraybackslash}m{0.3\columnwidth}|>{\arraybackslash}m{0.6\columnwidth}|c}
		\hline
			\rowcolor{DGray}
			
			\rowcolor{Gray} \multicolumn{3}{l}{\textbf{Topics}}   \\
			\hline    
			\textbf{Positively correlated} & most frequent words used by high-stress people & r  \\
			\hline    
			Exhaustion     & i'm, tired, hungry, bored, exhausted, freaking, sleepy, sooooooooo, stressed, grumpy           & .21 \\
			Feeling hurt      & i'm, sick, tired, feeling, hearing, tire, fed, bullshit, assuming, hurting, numb  & .21 \\
			Physical Pain & don't, feel, anymore, beg, begging, honestly, knees, clue, hollow, creeping, sympathy           & .20 \\
			Feeling   sick         & feel, sick, crap, feeling, ugh, hate, feels, sucks, crappy, bleh, worse, miserable, sickness, icky, =(        & .19 \\
			Losing control          & i'm, it's, i've, don't, lost, mind, quarter, wouldn't, anymore, control, reason 
			& .19 \\
			\hline
			\textbf{Negatively correlated} & most frequent words used by low-stress people & r  \\
			\hline    
			Family and eating      & great, lunch, nice, dinner, family, enjoyed, church, wonderful, afternoon, sunday, kids, evening, shopping, meeting, hubby
			& .13 \\
			
			Travel     & kings, delhi, leon, mumbai, rocks, queens, reached, rains, phew, bak, travelling, royal, metro, indians, mahal
			& .12 \\
			\hline
		\end{tabular}
	\label{t:stresstopics}
\end{table}

\textbf{Topics:} Topics which significantly correlate with stress are provided in Table \ref{t:stresstopics}. Exhaustion is typical of prolonged stress \cite{mcmanus2002causal}. Feeling hurt, physical pain, and feeling sick are known correlates of stress \cite{gil2004daily}. Lack of control also signifies the concept of resources falling short of demands, and possibly reducing their ability to cope with stressors \cite{gray2012moral}.

Our findings also reiterate previous research on language use in mental health \cite{de2013predicting}, as they describe symptoms of physical pain and sickness, besides expressing a lack of control and negative emotions. On the other hand, the mention of family meals on Facebook has a negative correlation with stress, suggesting a good social support network and taking time to spend with loved ones as well as travel, are ways people relax.  

\section{RQ2: Predicting User-Level Psychological Stress using Facebook and Twitter}

\subsection{Methods}

We utilize the Facebook and Twitter data from the same users, described in the previous section, to evaluate the performance of supervised models trained on Facebook and Twitter language at predicting users' stress for a held-out set. We stratify our set of users into five folds with a uniform distribution of age and gender traits in each fold.  We conduct a cross-validated weighted linear regression for stress, training on (a) LIWC features and (b) Topic features (c) TensiStrength scores, and (d) engagement features (such as time of posts, number of posts, number of posts between 12am-6am) for users in four folds, and testing on the users in the held out fold. 

In the five-fold cross-validation setting, we perform linear regression with several regularization methods such as ridge, elastic-net, LASSO and L2 penalized SVMs and find that elastic-net showed marginally superior performance over the others. Accordingly, we report results only using elastic-net. The performance was measured by calculating Pearson's $r$ over the aggregated predictions from the five folds. 

We first evaluate the above features to predict stress \textit{within domain} - i.e., we train and test on the same platform. Engagement features are part of the feature set used in \cite{lin2014user} for predicting user-level stress. We also examine how the models perform when compared to sociodemographic variables, namely age, gender, race, income, and education. 

Then, we evaluate how models trained on Facebook perform at predicting stress from Twitter language in a \textit{cross-domain} setting. Previous studies showed that predictive performance changes in cross-domain applications \cite{kokil2018,zhong2017wearing}. Therefore, we then attempt to improve the cross-domain prediction performance by applying \textit{domain adaptation}. The motivation for cross-domain and domain adaptation experiments is to build an accurate predictive model on Twitter language to then scale it to county level Twitter language. 

\subsection{Results for Within Domain Predictions}

\begin{table}[t!]
	\renewcommand{\arraystretch}{1.1}
	\setlength\tabcolsep{0.05in}
	\centering
		\caption{\small Within Domain: Stress Prediction Performance (Pearson's $r$) based on 5-fold cross-validation using different features and models trained and tested on the same domain. Social media language adds to and outperforms sociodemographic variables at predicting stress.}
	
	
	\begin{tabular}{l|c|c}
		\hline
		\rowcolor{DGray}

\rowcolor{Gray}		  \multicolumn{3}{c}{\textbf{D: Sociodemographic variables}} \\\hline
		\textbf{Feature}         &  \multicolumn{2}{c}{Pearson r} \\\hline
		 \begin{tabular}{@{}c@{}}Age, Gender, Race \\ Income \& Education\end{tabular}  &     \multicolumn{2}{c}{.25}                                                                 \\
		 \hline 
\rowcolor{DGray}

\rowcolor{Gray}		\multicolumn{3}{c}{\textbf{SM: Social media language}} \\\hline
		\textbf{Feature}         & \begin{tabular}[c]{@{}l@{}}Facebook\end{tabular} & \begin{tabular}[c]{@{}l@{}}Twitter\end{tabular} \\\hline
		\begin{tabular}{@{}c@{}}User Engagement \cite{lin2014user} \end{tabular} &     .11                                                                 &     .05                                                                 \\\hline
		\begin{tabular}{@{}c@{}}TensiStrength \cite{thelwall2017tensistrength} \end{tabular}  & .17 & .11 \\ \hline
		LIWC 2015      & .29                                                                 & \textbf{.22}                                                                \\\hline
		Topics         & \textbf{.31}                                                                 & .18            \\ \hline         
				\rowcolor{Gray} \multicolumn{3}{c}{ \textbf{Language + Sociodemographic}} \\\hline        
		\begin{tabular}{@{}c@{}} \textbf{D} + \textbf{SM} \end{tabular} & \textbf{.33} & .26 \\ \hline
		
	\end{tabular}
\label{withindomain}
\end{table}

Table \ref{withindomain} shows the performance of predicting stress using sociodemographic variables and social media language. Within the domain, Facebook does better than Twitter by a slight margin. On Facebook, both LIWC and Topics outperform sociodemographic variables (age, gender, race, income, and education). Topics outperform LIWC on Facebook (r=.305), and LIWC outperforms Topics on Twitter (r=.218). These correlations, which are considered a high correlation in measuring internal traits \cite{psychassessment}, show that linguistic features perform reliably well at predicting stress. User engagement features performed rather poorly when compared to linguistic features indicating that trait-prediction is a different task when compared to state-prediction. A similar observation has been made while using user engagement features for predicting other traits \cite{preotiuc2016studying}. The correlation between stress scores collected from survey and user-aggregated TensiStrength scores was .17 for Facebook and .11 for Twitter. TensiStrength was developed by annotating keyword-selected Twitter posts, which makes them inapplicable for measuring psychological stress. Consequently, we dropped engagement features and TensiStrength from subsequent analysis. We also limited the posting period to the previous one month (consistent with the time period in the Cohen's survey questionnaire) and observed that the performance shows a non-significant drop by $~0.02$ in $r$. 

\subsection{Results for Cross-Domain Predictions}
Since our goal is to predict stress in counties using Twitter, we examined how models trained on Facebook perform on Twitter (shown in Table \ref{crossdomain}). Performance drops (by ~5\% compared to within domain performance) when Facebook models are used to predict stress from the Twitter language. Specifically, topics see a larger drop (by ~50\%) possibly due to not being able to generalize across platforms. We have seen that expressions of stress in Facebook language vs. Twitter language are significantly different in terms of the vocabulary used. Consequently, a standardized theory-driven dictionary such as LIWC is stable across both platforms. We also combined both Facebook and Twitter corpora, and the model trained on both platforms together gives a marginal improvement in prediction performance.  

Given the marked distinction between expressions of stress in Facebook and Twitter language, we investigate several approaches to improve the predictive performance of models on Twitter. This problem can be viewed as a domain adaptation task, where we are adapting from a {\it source domain}: users' language on one platform, to a {\it target domain}: the same users' language on another platform.

\begin{table}[ht!]
	\renewcommand{\arraystretch}{1.1}
	\setlength\tabcolsep{0.02in}
	\centering
		\caption{Cross Domain: Stress Prediction Performance (Pearson's $r$) based on 5-fold cross-validation. FB: Facebook; Tw: Twitter.}
	
	\begin{tabular}{l|c|c}
		\hline
	\rowcolor{DGray}

\rowcolor{Gray}	\textbf{Feature}         & \begin{tabular}[c]{@{}l@{}}\textbf{Trained on FB} \\ \textbf{Tested on Tw}\end{tabular} & \begin{tabular}[c]{@{}l@{}}\textbf{Trained on FB+Tw} \\ \textbf{Tested on Tw}\end{tabular} \\\hline
		LIWC 2015       & .23                                                                 & \textbf{.24}                                                                 \\\hline
		Topics          & .15                                                                 & .17            \\ \hline                                      
	\end{tabular}
	\label{crossdomain}
\end{table}

\subsection{Results with Domain adaptation}
Most prior work on domain adaptation has focused on the case where some labels are available on both the source and target domains and is usually done by combining (often in some weighted fashion) training data sets or, less commonly, trained models from the source and target domains. A simple and effective supervised method was proposed by Daum\'e \cite{daume2009frustratingly} which applied a supervised heuristic mapping from labeled data the source and target domains, to a higher dimensional feature space, which are used to train standard classifiers or linear regression models. This approach has demonstrated the best performance in several comparative evaluations conducted for image-, text- and sentiment-classification \cite{pan2010survey}.

We test two standard domain adaptation techniques for improving the cross-platform performance of predictive models: one supervised approach; Easy Adapt \cite{daume2009frustratingly}, which uses labeled data from both the source and the target, and one unsupervised approach: Transfer Component Analysis -- TCA \cite{pan2011domain}, which requires no labels on the target domain. While there are several candidate algorithms within both supervised and unsupervised approaches, we chose Easy Adapt and TCA for their simplicity in application.

The standard notation used throughout this section is that $X_{S}$ refers to labeled observations in the source domain and $X_{T}$ refers to the test set in the target domain.

\paragraph{EasyAdapt}: We define our problem according to the implementation described by Daum\'e \cite{daume2009frustratingly}. Let $X1$ denote the original feature space for the user in the source domain, $X1 = \mathbb{R}^{F}$. We construct an augmented feature space $\widetilde{X1} = \mathbb{R}^{3F}$, by creating an platform-specific, and user-specific version of each feature in $X$. For this, we define $\Phi_{s} :X \longrightarrow \widetilde{X} $ to transform feature vectors corresponding to the platform-specific and user-specific feature spaces. The mappings are defined by the following equation: 
\begin{equation}
\small
\centering
\Phi^{s}(x) = \big \langle x,x,\mathbf{0}\big \rangle , \Phi^{t}(x) = \big \langle x, \mathbf{0},x\big \rangle 
\end{equation}
Here, $\Phi^{s}(x)$ is the feature vector for the source domain, $\Phi^{t}(x)$ is the corresponding vector for the target domain. $\mathbf{0} = \big \langle 0,0,...,0\big \rangle \in \mathbb{R}^{F}$ is the zero vector.

We thus augment the original feature space with the labeled data points (for the same set of individuals) from the target domain, excluding the held out sample for testing. Next, we train a regression model between this augmented feature space and stress scores of the users. We similarly transform the feature space of the held out sample before prediction. 

\paragraph{Transfer Component Analysis (TCA)}
TCA exploits the Maximum Mean Discrepancy Embedding
(MMDE) for comparing the distributions between the source and target domain, based on the Reproducing Kernel Hilbert Space (RKHS). The empirical estimate of the distance between $D_{S}$ and $D_{T}$, 
$Dist(D_{S},D_{T})$, is 
\begin{equation}
||\frac{1}{n_{1}}\sum_{i=1}^{n_{1}}{s}_{i} - \frac{1}{n_{2}}\sum_{i=1}^{n_{2}}{u}_{i} ||_{H}
\end{equation}
where $u$ and $s$ are individual observations in $D_{S}$ and $D_{U}$, H is a universal RKHS and $\phi$ : $X \longrightarrow  $H.

After applying domain adaptation (results in Table \ref{da}), performance increases by ~16\% when compared to using Facebook models alone to predict stress from Twitter language without any domain adaptation. TCA is seen to outperform EasyAdapt. 

\begin{table}[h!]
	\renewcommand{\arraystretch}{1.1}
	\setlength\tabcolsep{0.02in}
	\centering
	\caption{Domain Adaptation: Stress Prediction Performance (Pearson's $r$) based on 5-fold cross-validation, trained on Facebook and tested on Twitter.}
	
	
	\begin{tabular}{c|c|c|c}
		\hline
		\rowcolor{Gray}
		\textbf{Feature} &    \begin{tabular}[c]{@{}l@{}}\textbf{No Domain}\\ \textbf{Adaptation}\end{tabular}    & \begin{tabular}[c]{@{}l@{}} \textbf{EasyAdapt}\end{tabular} & \begin{tabular}[c]{@{}l@{}}\textbf{Transfer Component}\\ \textbf{Analysis}\end{tabular} \\\hline
		LIWC 2015      & .23 & .23 & \textbf{.27} \\\hline
		Topics         & .15 & .16 & .18 \\ \hline
	\end{tabular}
	\label{da}
	
\end{table}

\section{RQ3 Evaluating County Level Language-Predicted Stress}

\subsection{Methods}

\subsubsection{Data} The language of US counties is obtained from Twitter. It comprises of the normalized frequency distributions of words, averaged by the number of individuals who spoke those words, extracted from a set of geo-located tweets \cite{schwartz2013characterizing}. The authors constructed this dataset by collecting a 10\% sample of Twitter posts between the years 2009-15 from the TrendMiner project \cite{preotiuc2012trendminer}. Tweets were geo-mapped to US counties using latitude/longitude coordinates and the self-reported location field when available, following the approach described in \cite{schwartz2013characterizing}. In this manner, ~20\% of the tweets could be successfully mapped, resulting in over 1.7 billion geo-located tweets. A threshold of 100,000 words per county is used in order to ensure that any word-to-outcome correlations observed were stable, resulting in a dataset of 2710 counties. On an average, each county had 8,892,568 words. 

\textbf{County Health outcomes:} We use two sets of county health outcomes in our work: 1. Gallup-Sharecare Well-Being data and 2. County health statistics for the United States. 

From the Gallup-Sharecare Well-Being Index, we use the stress outcome to validate the stress predictions made by our language model. Gallup data is collected as a part of 1,000 telephone interviews conducted every day across the US \cite{gallup}. The questions on the interview range from topics such as health behaviors, work environment to social and county factors, and financial security. We specifically used the stress fields from the Gallup data aggregated to counties.

From US County Health Rankings and Roadmaps portal\footnote{\url{www.countyhealthrankings.org/}}, we obtained county socioeconomic characteristics and health behaviors which provides access to county-level health factors from a wide range of sources, including Behavioral Risk Factor Surveillance System, American counties Survey, and the National Center for Health Statistics. 

\subsection{Results}

\begin{figure}[h!]
	\centering
	\includegraphics[width=1\columnwidth]{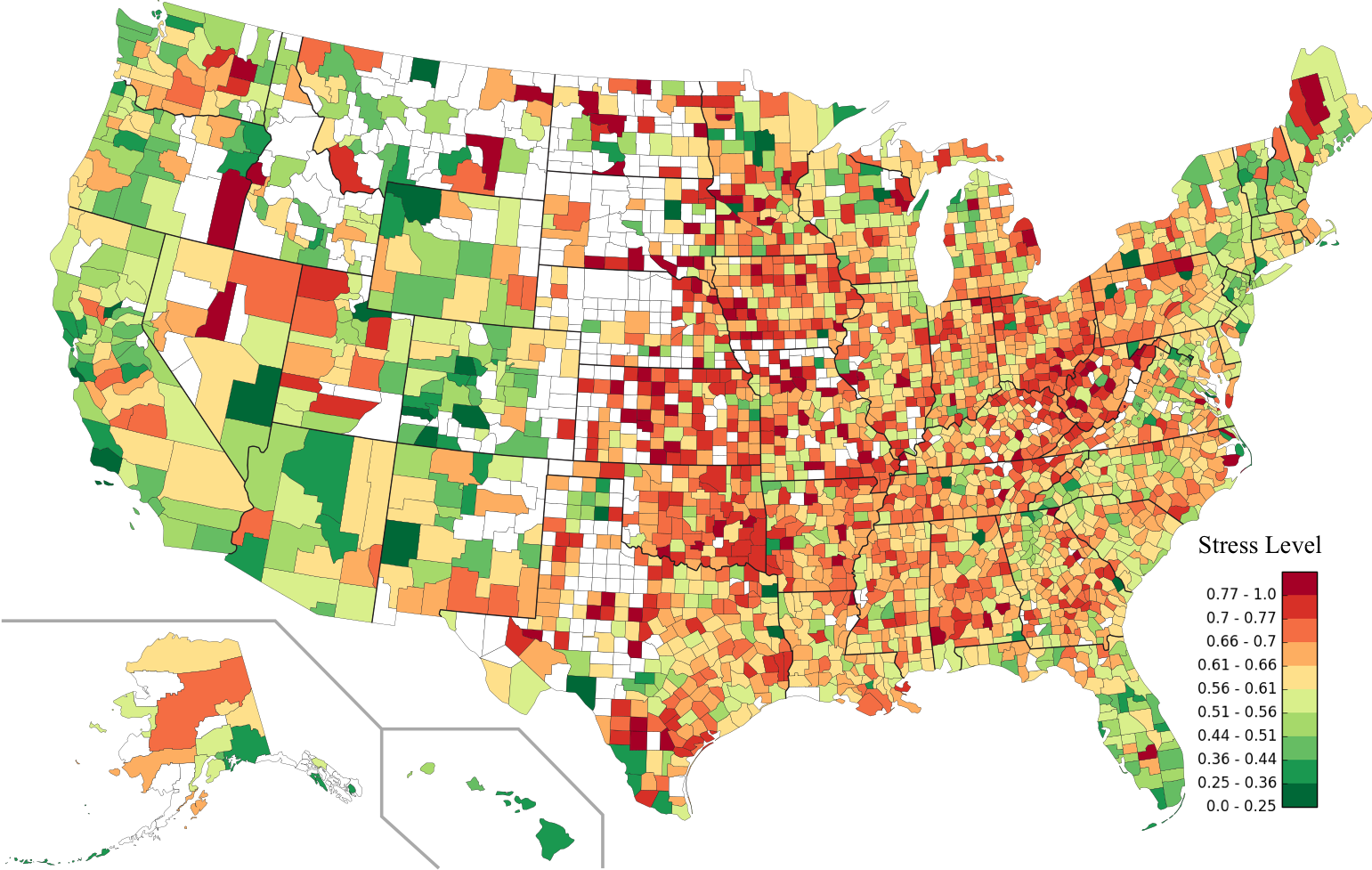}
	\caption{Map showing stress predicted from Twitter language from US counties. Counties which had less than 100k tweets are shown white.}
	\label{fig:counties}
\end{figure}

After analyzing the performance of different models at predicting stress in the cross-domain setting, we use the domain-adapted model trained on LIWC features to predict stress using Twitter language from US counties (shown in Figure \ref{fig:counties}). We validate our language-predicted stress with county-level stress reported by Gallup, and further with county-level estimates of health behaviors and socioeconomic characteristics.

First, we use the Gallup-Sharecare Well-Being data, with county-aggregated stress labels (termed as `Gallup stress' hereon), as a means of validating our language-predicted stress. We compare Gallup stress and language-predicted stress (referred to as `Twitter stress' hereon). We aim to train the best performing user-level model, as this directly impacts the county-level prediction accuracy. We experiment with models trained on individual platforms (best $r = 0.26$), and find a significant difference in predicting Gallup stress after domain adaptation (r = $0.34$). These correlations are considered a high correlation in psychology, especially when measuring outcomes like stress \cite{psychassessment}. Part of this improvement is due to the amount of information that Facebook contains over Twitter (Table \ref{withindomain}). However, Facebook alone is insufficient as the Twitter language is substantially different from Facebook (Tables \ref{crossdomain} and \ref{da}). 

Predictive performance of sociodemographic variables (age, gender, income and education) and Twitter language are r=.24 and r=.34 respectivelt. Twitter language outperforms sociodemographic variables at predicting stress in counties.

Having validated the language-estimated stress in counties, we examine the correlation of stress with socioeconomic characteristics and health behaviors in US counties. The results are shown in Table \ref{t:uscounties}.  

\paragraph{Health Behaviors}

Counties with higher stress have higher mortality, a higher percentage of physically inactive people, a higher percentage of poor mental health days, and a higher percentage of smokers, teen pregnancy, and higher rates of drug poisoning. These are all factors which have known correlation and causality structures as reported by \cite{stults2014effects,steptoe1996stress,watson1989health}. Individuals living in high-stress counties appear to suffer from poor health, affected both by aspects in their individual lives as well as by the nature of their counties. Using social media language as a real-time proxy to assess a county's stress, and consequently, health, could help design interventions and policies to alleviate stressors. Language has the specific advantage of getting at the causes/stressors that counties are experiencing, and this could greatly complement and also inform traditional survey mechanisms (such as Gallup).

\paragraph{Socioeconomic Characteristics}
Counties which are rich and educated express less stress \cite{karlamangla2006reduction}. This is observed by the low correlation between language-predicted stress and Median Household Income (r=-.271) and Education level, \% people who attended some college, (r=.345) in counties. Furthermore, counties where people have access to exercise facilities also have low stress \cite{salmon2001effects}. Our language-based findings of stress and income are corroborated by an analysis of the Gallup data reported by \cite{de2017does}.  High-stress counties also have higher avoidable hospitalization rates and have less access to providers for mental and physical health. These associations suggest that individuals in high-stress counties also face other economic and infrastructural challenges. 

\begin{table}[t!]
	\centering
	\footnotesize
		\caption{Correlation between county level health measures and stress. All correlations are significant at $p<0.01$. Language-based stress predictions predict health behaviors and clinical care facilities above and beyond socio-demographic controls}
	\begin{tabular}{L{4cm}cc}
		\hline 
	\rowcolor{DGray}
 \hline                    
\rowcolor{Gray}	\textbf{County-level Outcomes}                                                                        & \multicolumn{2}{c}{\textbf{Language Correlation}}            \\
	\rowcolor{DGray}
 \hline                    
\rowcolor{Gray}		& \begin{tabular}[c]{@{}l@{}} With \\No Controls\end{tabular} & \begin{tabular}[c]{@{}l@{}}With SES\\ controls\end{tabular} \\
		\hline

		\multicolumn{3}{l}{\textbf{Health Behaviors}}                                                                                                                               \\ \hline
		MV Mortality Rate                                                                                     & .40                & .11                     \\
        \% Physically Inactive                                                                                & .44                & .31                     \\
		\% Smokers                                                                                            & .29                & .30                     \\
        \% Poor mental health days                                                                            & .14                 & .13                     \\

		Teen Birth Rate                                                                                       & .23                & -                                            \\
		
		Drug Poisoning Mortality Rate                                                                         & .13                & -                                            \\
	\hline
		\multicolumn{3}{l}{\textbf{Clinical Care}}                                                                                                                                  \\ \hline
	Preventable Hosp. Rate                                                                                & .41                & .36                     \\

	Mental health providers rate                                                                          & -.39               & -.19                     \\
	Primary Care Physician Rate                                                                           & -.43               & -.33                     \\ 
		\% Could not see doctor due to cost                                                                   & .17                 & -   \\\hline
	
			\multicolumn{3}{l}{\textbf{Socioeconomic Environment}}                                                                                                                     \\ \hline
	Median Household Income (ln) & -.27 & - \\
	Education (\% College graduates) & -.34 & -  \\
		\% With Access to Exercise Facilities                                                                 & -.45               & -.264                      \\
	
 \hline

	\end{tabular}%

	\label{t:uscounties}
\end{table}

\section{Discussion}
Our work provides some of the first insights into the linguistic manifestation of psychological stress on social media, while also showing the feasibility of applying models trained at the user-level to predict stress in counties with validity.  Language associated with stress is indicative of exhaustion \cite{mcmanus2002causal}, comorbidity with mental health conditions \cite{watson1989health}, pronounced self-focus \cite{pennebaker2002language} and feelings of hurt, physical pain, and being sick, which also known correlates of stress \cite{gil2004daily}. Our findings also reiterate previous research on language use in mental health \cite{de2013predicting}. An important insight from our work was that the predictive utility of different platforms varies, as has been seen in prior research looking at differences in self-disclosure \cite{kokil2018,zhong2017wearing}. This motivates the need for transfer learning when the stress model trained on Facebook language needs to be applied to Twitter language. This observation has implications for other researchers interested in training and applying machine learning models across domains. 
 
On applying the adapted stress model to the language of counties, we obtain face-valid results about the relationships between stress and ground-truth statistics about county health. Our findings clarify the association of stress with socioeconomic factors \cite{baum1999socioeconomic}, suggesting that it is the socially and economically deprived counties which are likely to face higher stress. Urban counties, with pressures of work and a fast-paced routine, are less likely to experience debilitating, psychological stress \cite{lederbogen2011city}. This is an important way to differentiate the stressful events in an urban lifestyle with the trait-based stress experienced in deprived or rural counties.

\subsection{Limitations and Future Work}
While there are several advantages and merits of using social media to measure stress, one of the apparent shortcomings of linguistic analysis is that they do not offer causal insights. Our methods anticipate that the language of counties is indicative of stress. It is possible that people might stop using social media while they are stressed, which could impact the performance of our user-level model \cite{corrigan2001familiarity}. However, we did not find any correlation (positive or negative) between the raw number of posts and psychological stress. To address such scenarios, monitoring multiple data sources such as phone-based sensors can be useful in promptly reaching out to the afflicted individuals \cite{singh2016cooperative}. Further, it would be interesting to disentangle the language indicative of short-term stressors versus long-term chronic stress, potentially by collecting data around recent stressful events (divorce, moving house, losing job etc.).

Some studies have shown preliminary findings that connect social media usage to poorer mental health, indicating that undesirable usage of social media might cause depression among its users \cite{o2011impact}. While we can see that social media can uncover several behavioral and psychological insights into the stressful lifestyles of people, further research is warranted to precisely determine the causality pathways.

While one way of representing counties is by their geographic distribution (as in the case of US counties), another way could be by their residents' occupation (`work' counties), and students (`college' counties). Monitoring stress levels in such counties would especially be useful in devising programs or initiatives to encourage a low-stress lifestyle amongst wherever individuals are particularly vulnerable to stress. While social media (and Twitter in particular) are not representative of the real-life users, several insights have been obtained using public Twitter data in public health \cite{yang2018retweet}. Further, the underlying predictive model can be applied to several online forums. For example, it could be applied to colleges across the U.S. to understand the specific chronic stressors in different college campuses (extending the recent work \cite{saha2017modeling,bagroy2017social}). It is especially promising that such measurements can be effective while being unobtrusive. Monitoring stress using social media also offers the advantage of discovering stress associations that are otherwise not easy to get by traditional surveys \cite{holdeman2009invisible}. Moreover, such a tool could monitor stress in real time, potentially detecting unforeseen events before they happen \cite{reason1995understanding}.  

On the technical front, while our application of domain adaptation bridged the gap between two social media platforms -- Facebook and Twitter, we note that further work is warranted to address the challenges associated with spatially correlated terms introduced when aggregating tweets to counties. The study by \cite{rieman2017domain} has explored such problems and their implications for county-level predictions from language. Also, several works that focus on socio-geographical studies have used advanced autoregressive modeling techniques which could help in developing more robust models, even for counties with missing data based on their spatial neighborhood association with other counties that have data points. Further, it would be interesting to examine stress from multiple modalities (text, images and sensor data), which can potentially complementary insights \cite{sharma2012objective,guntuku2015evaluating,samani2018cross}, particularly transfering learning across modalities~\cite{guntuku2019imagedep}. 

\subsection{Privacy and Ethical Considerations}
Prediction performances obtained in this study indicate that psychological stress can be inferred with some accuracy from public (Twitter) or semi-public (Facebook) social media data. While the purpose of delivering social support and mental health services  motivates this effort, these algorithms also raise some important privacy and ethical questions. 

From the perspective of privacy concerns, organizations with vested interests (e.g. insurance companies) may be motivated to infer this information automatically. As being chronically stressed can lead to stigma at multiple levels, data protection and ownership frameworks are necessary to make sure the data is not used against the users’ interest \cite{mckee2013ethical}. Few users realize the amount of health-related information that can be inferred from their digital traces; consequently, transparency about the indicators derived and the purposes behind such inference should be part of ethical and policy discourse. 

\section{Conclusion}
Using a sample of users who have accounts on both Twitter and Facebook, we uncovered insight into the language of stress in individuals. Our results also showed that standardized linguistic dictionaries, such as LIWC, outperform engagement attributes such as user posting behavior, in predicting stress. Our results complement psychological survey data by deepening our understanding of the environment that may contribute to both individual and county-level well-being. Language estimates can be used to monitor stress in different social settings, to devise initiatives encouraging a low-stress lifestyle. Our model will help measure the impact of such interventions in real time. Further, language has the specific advantage of getting at the causes/stressors that counties are experiencing thereby enabling personalized interventions.

The findings shed light on the nuanced relationship of socioeconomic status with psychological stress and suggest insights for other psychological factors, such as locus of control, the degree to which people feel like being manipulated by their environment vs. perceive a sufficient level of self-control and in the counties where they live and work. We believe that the potential for this technology to be used in both micro and macro scales is tremendous. Our computational techniques, while useful to monitor stress in counties, can also be used to give real-time feedback to individuals in those counties. Such feedback about individuals' social media usage along specific dimensions, can help in designing simple but effective technology-assisted interventions in cultivating mindfulness and stress control \cite{miller1995three}.  

\bibliographystyle{aaai}
\bibliography{FULL-GuntukuS.52} 
\section{Appendix}
Here, we describe the experiments detailing the language differences between Facebook and Twitter and their associations with stress.    

\begin{figure*}[t!]
      \centering
      \includegraphics[width=1\textwidth]{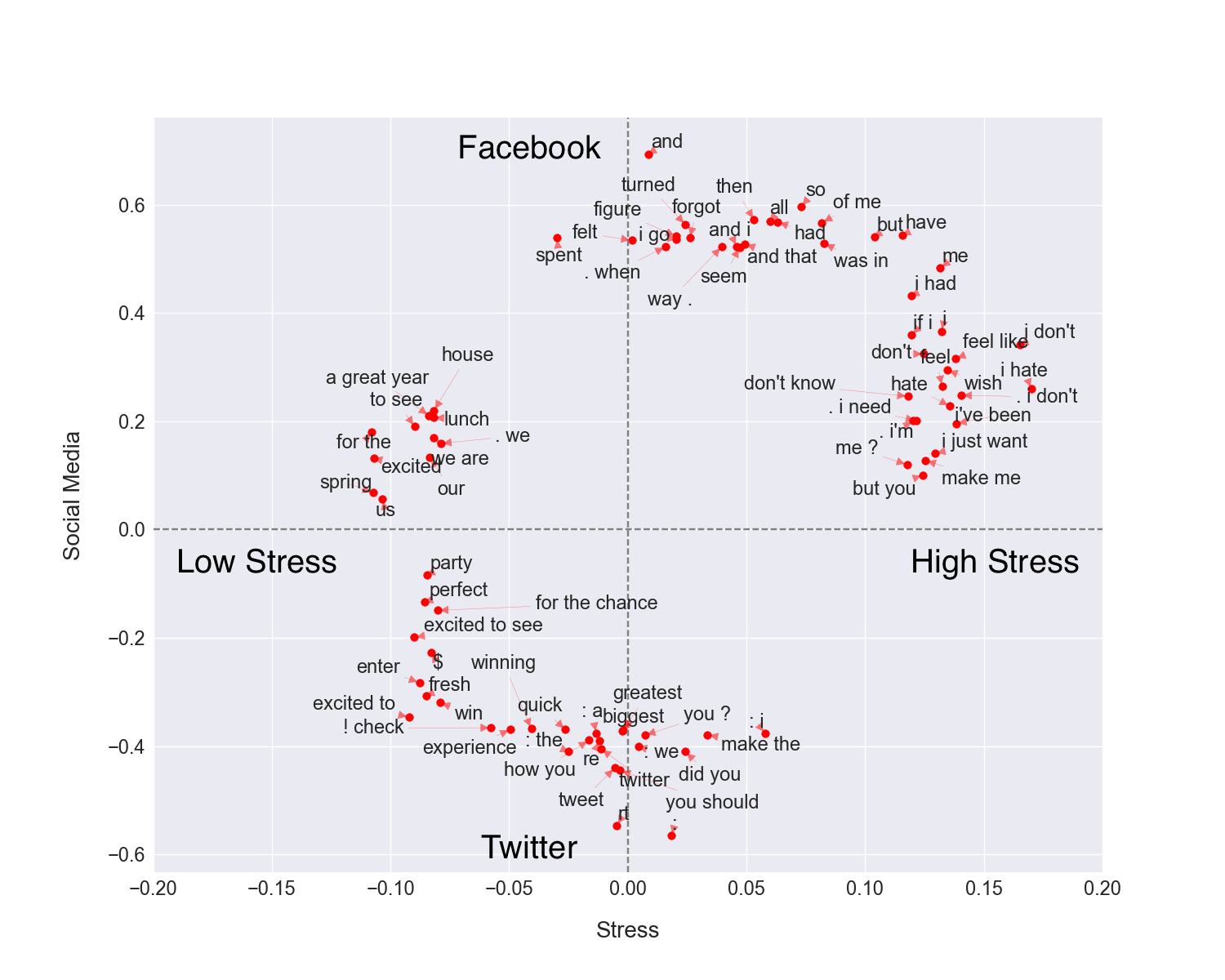}
      \caption{Correlations of words/phrases with Social Media platform (Facebook on the positive y-axis and Twitter on the negative y-axis) and Stress Scale. Correlations are significant, Bonferroni corrected at $p<0.01$}
\label{fig:stress_2d_ngram}
\end{figure*}

\paragraph{Usage Statistics:} In terms of pure usage statistics, we find that the our 601 participants had an average of 1238 Facebook and 1649 Twitter posts. The mean number of words they wrote per post was 17.1 for Twitter and 23.8 for Facebook. This difference is expected given the 140-character limit imposed on Twitter posts. However, the {\it median} number of words per post -- 17 for Twitter and 18  for Facebook -- is much closer, suggesting that our participants were more varied in their posting behavior, when it came to Facebook.

\paragraph{N Grams:} Next, we consider platform differences in expressing stress, at the word- and phrase-level. Figure \ref{fig:stress_2d_ngram} provides a 2D visualization of the Spearman correlations of individual words with stress on the X-axis, and with social media platform on the Y-axis. The words along the X-axis are highly correlated with stress, but less preferential towards individual social media platforms. The words at the Y-axis poles are highly correlated with platform usage but less predictive of stress. For the purpose of this plot, we have discarded words with non-significant correlations as well as $\rho <= 0.05$. Thus, the words along the diagonals are the ones which best distinguish Twitter usage from Facebook usage for the purpose of expressing stress. 

On Facebook, words and phrases such as ``me'', ``i had'', ``feel like'', ``i don't'' and ``i hate'' are significantly correlated with high stress, while words and phrases such as ``lunch'',``a great year'' and ``to see'' are significantly correlated with low stress. On the other hand, on Twitter, words and phrases which indicate low stress are ``excited to'', ``! check'', ``fresh'', ``win'' and ``experience''. 

The lack of words in the bottom right quadrant suggests that that language of high stress on Twitter is a subset of the language of high stress on Facebook. These findings motivate our proposal to adapt stress models trained on Facebook language, a more inclusive corpus, to Twitter language, for the purpose of county-level stress prediction.

\paragraph{LIWC:} From Table \ref{t:liwc_tw}, we observe that Twitter has some similarities with Facebook - such as the use of adverbs, and the negative correlation with the use of ``Positive Emotion'' words. Unexpectedly, on Twitter, the category most strongly correlated with stress is ``Comparisons'' and not first person singular pronouns. We anticipate that this reflects the tendency of Twitter users to drop self-referencing pronouns, as has been identified in previous work \cite{rouhizadehusing}. Comparisons and differentiation could indicate lack of satisfaction with status quo, indicating possibly pessimism or expressing perceived resources to fall short of demands, or simply lower perceived satisfaction with life of those prone to stress~\cite{linn1985health}. Insecurities of those higher in stress may be what is indicated by correlations with nonfluencies~\cite{felson1993review}. A ``focus on the past'' tends to be maladaptive and may indicate that those high in stress might find themselves stuck in what happened and in ruminative, self-deprecating thoughts. This is reinforced by the negatively correlated category ``Power'', which suggests that the individuals under stress express a lack of control and superiority over their environment~\cite{abouserie1994sources}.

\paragraph{Topics:} Twitter topics which are positively correlated with stress (Table \ref{t:stresstopics_tw}) comprise of words and emojis depicting negative emotions, possible sarcasm and feelings of awkwardness. We did not find any topics which are negatively correlated with stress, on Twitter when controlled for age and gender.

\paragraph{TensiStrength:} The Pearson correlation between the mean stress values for Facebook and Twitter, was 0.28 and significant at $p<0.01$. This difference is expected because the TensiStrength lexicon was developed on Twitter posts, and as such it might not generalize well to Facebook language. The finding does suggest that stress is expressed very differently on Twitter versus Facebook, and motivates our decision to adapt Facebook stress models to Twitter.

\begin{table}[t!]
\renewcommand{\arraystretch}{1.1}
\setlength\tabcolsep{0.02in}
\centering
\footnotesize
\resizebox{\columnwidth}{!}{
\begin{tabular}{ >{\centering\arraybackslash}m{0.3\columnwidth}|>{\arraybackslash}m{0.45\columnwidth}|c}
\rowcolor{DGray}

\rowcolor{Gray} \multicolumn{3}{l}{\textbf{LIWC - Twitter}}   \\
\hline    
\textbf{Positively correlated} & most frequent words & r  \\
\hline    
Comparisons    & greater, best, after   & .164 \\
Adverbs       & very, really & .144 \\
Nonfluencies & er, hm, umm    & .136 \\
Differentiation & hasn't, but, else    & .134 \\
Focused on Past & ago, did, talked    & .134 \\

\hline
\textbf{Negatively correlated} & most frequent words & r  \\
\hline    
Power       & superior  & -.159 \\
Positive Emotion      & love, nice, sweet   & -.107 \\
\hline
\end{tabular}
}
\caption{Pearson correlations (maximum 5) between the stress score and LIWC features extracted on Twitter, when controlling for age and gender. All features are significant at $p<.01$, two tailed t-test.}
\label{t:liwc_tw}
\end{table}

\begin{table}[t!]
\renewcommand{\arraystretch}{1.1}
\setlength\tabcolsep{0.02in}
\centering
\footnotesize
\resizebox{\columnwidth}{!}{
\begin{tabular}{ >{\centering\arraybackslash}m{0.2\columnwidth}|>{\arraybackslash}m{0.65\columnwidth}|c}
\rowcolor{DGray}

\rowcolor{Gray} \multicolumn{3}{l}{\textbf{Topics -- Twitter}}   \\
\hline    
\textbf{Positively correlated} & most frequent words used by high-stress people & r  \\
\hline    
 Hatred     & hate, stupid, $-_-$, $>.<$, gah, ugh, friggin, effin, sigh, grr, $>:$,wtf, annoying, urgh, dammit & .165 \\
 Sarcasm      & funny, it's, thinks, people, isn't, finds, hilarious, ironic, entertaining, nicer, unfortunate & .159 \\
 Feeling   awkward         & feels, weird, kinda, weird, bit, strange, odd, hmm, suddenly, sort, awkward, dunno        & .150 \\
\hline
\end{tabular}
}
\caption{Pearson correlations between the stress score and Topics extracted from Twitter posts, when controlling for age and gender. Topic labels are manually created. All features are significant at $p<.01$, two tailed t-test. No negative correlations were found for Twitter posts}
\label{t:stresstopics_tw}

\end{table}

\end{document}